\documentclass{article}
\usepackage{spconf,amsmath,graphicx,subfigure,algorithm,algpseudocode,booktabs,caption,flushend}
\usepackage{tabularx}

\title{SINGLE ARCHITECTURE AND MULTIPLE TASK DEEP NEURAL NETWORK FOR ALTERED FINGERPRINT ANALYSIS}

\name{Oliver Giudice$^{\star}$, Mattia Litrico$^{\star \dagger}$, and Sebastiano Battiato$^{\star}$}

\address{$^{\star}$ Department of Mathematics and Computer Science, University of Catania, Catania, Italy\\
$^{\dagger}$iCTLab s.r.l. - Spinoff of University of Catania, Catania, Italy\\
\emph{\{giudice,battiato\}@dmi.unict.it, mattia.litrico@studium.unict.it}
}

\begin{document}

\maketitle

\begin{abstract}
Fingerprints are one of the most copious evidence in a crime scene and, for this reason, they are frequently used by law enforcement for identification of individuals. But fingerprints can be altered. "Altered fingerprints", refers to intentionally damage of the friction ridge pattern and they are often used by smart criminals in hope to evade law enforcement. We use a deep neural network approach training an Inception-v3 architecture. This paper proposes a method for detection of altered fingerprints, identification of types of alterations and recognition of gender, hand and fingers. We also produce activation maps that show which part of a fingerprint the neural network has focused on, in order to detect where alterations are positioned. The proposed approach achieves an accuracy of 98.21\%, 98.46\%, 92.52\%, 97.53\% and 92,18\% for the classification of fakeness, alterations, gender, hand and fingers, respectively on the SO.CO.FING. dataset.
\end{abstract}


\section{Introduction}\label{sec:intro}

Fingerprints have long been used to identify individuals \cite{Meuwly2009}. Thus are one of the most accurate and numerous biometric data in crime scenes representing the most valuable information for investigators. In addition, security agencies in some countries have developed, since 1999, the Automated Fingerprint Identification Systems (AFIS) \cite{Meuwly2009}, a system capable of verifying the identity of a subject based on his fingerprints and employed, among other things, for the identification of criminals who attempt to enter foreign territory. With intact and high quality fingerprints, AFIS is able to achieve high levels of accuracy and robustness but performances degrade greatly when low quality, altered or incomplete fingerprint images are involved. For instance, the destruction, intentional or not, of the dermatoglyph crests present in fingers, strongly degrades the information contained in the image, greatly increasing the percentage of errors committed by the AFIS system in the identification process.
State of the art in the image forensics field such those described in \cite{battiato2016multimedia}, \cite{piva2013}, \cite{stemm2013} even the newest CNN-based ones like \cite{kuznetsov2019new} are not the best solution: here integrity and authenticity of images is not compromised while being the produce image imprint the effect of a voluntarily process of degradation of information of the actual fingerprint structure itself \cite{tabassi2018altered}.

Fingerprint alteration is not a new issue for investigators: the first altered fingerprint case dates back to 1896, when Sir Francis Galton, a British explorer, anthropologist and climatologist, documents the first transplantation of the skin of a fingertip.
Galton told that, in an attempt to cut cardboard with a sharp knife, injured his finger by inadvertently severing a portion of skin from his fingertip. Subsequently, the man managed to recover the cut epidermis layer, and applied it again on the wound by immobilizing it with bandage. Eventually, the transplant was successful and even 30 years later, it is possible to recognize the alteration of fingerprint pattern: the flow of the dermatoglyphs was rotated at a right angle with respect to the original direction \cite{cummins1934attempts}.
In 1933, altered fingerprints were found in Gus Winkler, an American gangster, through cuts and lacerations, in all the fingers of his left hand, except for the thumb. Furthermore, the pattern of the dermatoglyphs in one finger was altered in almost all its entirety.
More recently, in 1995, a man named Alexander Guzman was arrested in Florida for having a fake passport. Guzman had mutilated his fingers by first making a "Z" cut (Z-cut) on the fingertips and, later, exchanging between them the skin that constituted the two triangles that had originated from the previously made cut. Only after two weeks of work, the FBI managed to manually reconstruct the characteristics of the original fingerprints and proceeded to identification, discovering that the real identity of Guzman was that of José Izquiredo, a drug dealer \cite{cummins1934attempts}.
Finally, in 2009, a Chinese woman managed to bypass security measures and illegally enter Japan thanks to a plastic surgery operation through which she had exchanged the prints of her right hand with those of the left \cite{tabassi2018altered}.

In this paper, a deep neural network approach based on Inception \cite{szegedy2016rethinking} will be presented able to deal with all the before-mentioned alterations. The objective was to build a \emph{single architetcture/multiple task} solution, to this aim the CNN was trained and tested on SO.CO.FING. dataset \cite{shehu2018sokoto}. Results demonstrated to over-perform state of the art techniques.

The remainder of this paper is organized as follows. Section \ref{sec:alt} will introduced the reader to the three alteration types that will be predicted by proposed approaches. In Section \ref{sec:related} the state of the art will be presented and commented. Section \ref{sec:approach} the classification solution will be described with details of training process and parameters employed and in Section \ref{sec:results} test for the five different tasks will be presented. Finally \ref{sec:conclusion} ends this paper.

\section{Alteration Types}\label{sec:alt}
\begin{figure}[t]
     \begin{subfigure}[]
         \centering
         \includegraphics[width=0.15\textwidth]{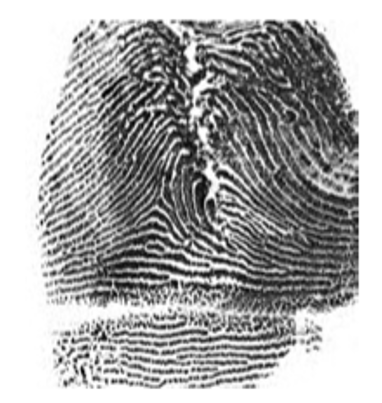}
     \end{subfigure}
     \hfill
     \begin{subfigure}[]
         \centering
         \includegraphics[width=0.15\textwidth]{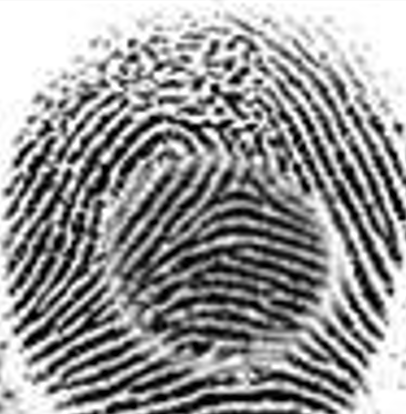}
     \end{subfigure}
     \hfill
     \begin{subfigure}[]
         \centering
         \includegraphics[width=0.15\textwidth]{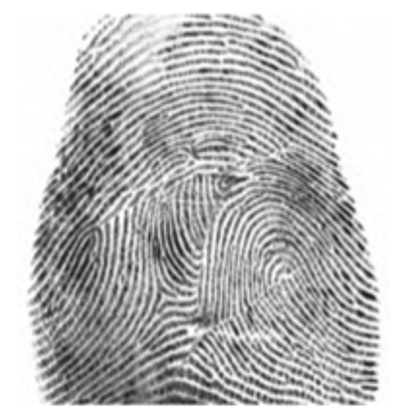}
     \end{subfigure}
        
        \caption{Example of fingerprint alteration types: (a) Obliteration, (b) Central Rotation and (c) Z-cut.}
        
        \label{fig:alteration_examples}
        
\end{figure}

Alteration types are defined on the basis of how the flow of dermatoglyph crests changed after the alteration itself. Thus, altered fingerprints were classified by Yoon et al. \cite{yoon2012altered} as: obliteration and distortion. In particular distortion can be specified into central rotation and z-cut. Figure \ref{fig:alteration_examples} show examples of each kind of alteration. These alterations can be defined as follows:

\begin{itemize}
    \item \emph{Obliteration}: is the most common type of alteration. This is due to the simplicity of employing this process w.r.t. the distortion produced. On the other hand, however, the detection of an obliterated impression is much simpler than the other types. The obliteration can be carried out by abrasion, cuts, burns or application of chemicals.
    \item \emph{Distortion}: The distortion of a fingerprint causes the presence of an unnatural pattern. The anomalies in the ridges pattern are referred to: an abnormal spatial distribution of the singular points or of the regions of the fingerprint in which the curvature of the ridges is greater than normal and their direction changes rapidly \cite{bo2009novel}. In other words, a sudden change of orientation of the ridges along the scar produced by the alteration. Despite this, a distorted impression can still circumvent the tests that are carried out on it, since, in some cases, its local properties, such as the trend of the dermatoglyph crests, or global, such as the length and frequency of the crests, remain unchanged. The distortion can be done by removing regions of skin from the fingertips and replanting them in a different position. The way the replanting is done, defines two different kinds of distortion. The first is the central rotation, that appears when the patch, even from other fingers, is re-planted in place with a rotation. The second is the Z-cut, that is caused by a more complex process consisting of three steps: making a ‘Z’ shaped cut on the fingertip, lifting and switching two triangular skin patches, and stitching them back together.
\end{itemize}

\begin{figure}[t]
     \begin{subfigure}
     \centering
         \includegraphics[width=0.15\textwidth]{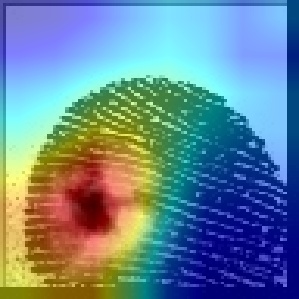}
     \end{subfigure}
     \hfill
     \begin{subfigure}
     \centering
         \includegraphics[width=0.15\textwidth]{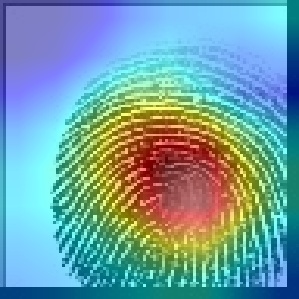}
     \end{subfigure}
     \hfill
     \begin{subfigure}
     \centering
         \includegraphics[width=0.15\textwidth]{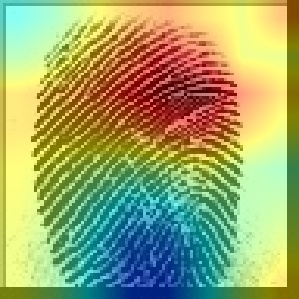}
     \end{subfigure}
     \begin{subfigure}
     \centering
         \includegraphics[width=0.15\textwidth]{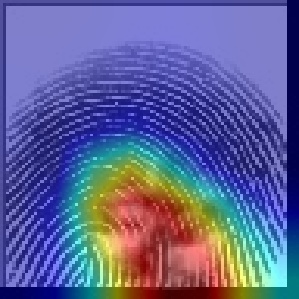}
     \end{subfigure}
     \hfill
     \begin{subfigure}
     \centering
         \includegraphics[width=0.15\textwidth]{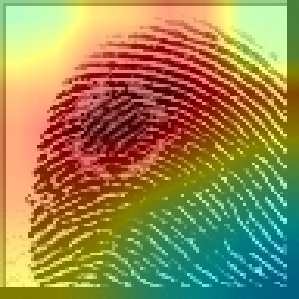}
     \end{subfigure}
     \hfill
     \begin{subfigure}
     \centering
         \includegraphics[width=0.15\textwidth]{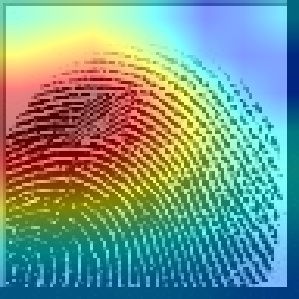}
     \end{subfigure}
     \caption{Activation maps of different types of alterations: (a) and (d) Obliteration; (b) and (e) Central Rotation; (c) and (f) Z-cut}
     \label{fig:activationMaps}
\end{figure}

\section{Related Works}\label{sec:related}

The classification of altered fingerprints has been studied by many researches through years, using a wide variety of techniques and also aimed at achieving multiple and different classification objectives.

Feng et al. \cite{feng2010detecting} trained a Support-Vector-Machine (SVM) to locate irregularities in the orientation of the flow of the dermatoglyph ridges. Evaluating their model on a dataset consisting of 1,976 simulated fingerprints, they reported a detection rate of 92 \% against a 7 \% of false positives. 

Ellingsgaard and Busch \cite{ellingsgaard2014detecting} \cite{ellingsgaard2017altered}, discussed a method that allows to automatically detect altered fingerprints based on the analysis of two different local characteristics of the image: the identification of irregularities in the pixel-wise orientation or the orientation of the signal in every single pixel and the analysis of the orientation of the minutiae in local patches.

Jain et al. \cite{tabassi2018altered} have trained two neural networks, specifically "Inception-v3" and "Mobile-net", obtaining a True Detection Rate (TDR) of 99.24 \% and ~ 92 \% respectively against a False Detection Rate of 2 \%. For this purpose, they used a dataset consisting of 4,815 altered fingerprints from 270 subjects and acquired by law enforcement and additional 4,815 valid rolled fingerprints, i.e. fingerprints acquired by rolling the finger in the appropriate paper, previously properly prepared.

Shehu et al. \cite{shehu2018detection}, using the "SO.CO.FING." or simply SOCOFING dataset \cite{shehu2018sokoto}, employed a Convolutional Neural Network (CNN) in order to discriminate valid fingerprints from the altered ones and, if present, to classify the type of alteration according to three types: obliteration, central rotation and Z-cut. Their approach obtained an accuracy of 98.55 \%.

Shehu et al. again in \cite{shehu2018detailed}, proposed a Transfer Learning solution, based on the CNN of the previous study, for classification of the fingerprints in the "SOCOFING" dataset in relation to the sex of the individual to which they belong, obtaining an accuracy of 75.2 \%.

State of the art shows the effectiveness of CNN-based approaches for this problem, to this aim a \emph{single architecture and multi-task} solution will be proposed that demonstrated to improve state of the art in each of the before-mentioned tasks.

\begin{table}[t]
\setlength{\tabcolsep}{25.5pt}
\caption{Classification results (Fakeness). \textbf{Accuracy}: $98.21$\%, \textbf{Precision}: $99.27$\%, \textbf{Recall}: $98.16$\%} 

\begin{tabular}[t]{lccc} \toprule

\label{tab:fakeness}
& \textbf{Altered} & \textbf{Real} \\ \midrule
\midrule
\textbf{Altered}     & 43.970           & 820           \\
\textbf{Real}        & 9                & 1492         \\ \toprule
\end{tabular}

\end{table}

\begin{table}[t]
\setlength{\tabcolsep}{13.5pt}


\caption{Classification results w.r.t. alterations. \textbf{Accuracy}: $98.46$\%}
\begin{tabular}[t]{lccccc} \toprule
\label{tab:alteration}
& \textbf{Obl} & \textbf{Cr} & \textbf{Z-cut} & \textbf{Real} \\ \midrule \midrule
\textbf{Obl}     & 4460          & 1 & 0 & 6           \\
\textbf{Cr}        & 33                & 4019 & 19 & 27 \\
\textbf{Z-cut}     & 13           & 24 & 3695 & 13        \\   
\textbf{Real}     & 3           & 4 & 69 & 1424           \\\toprule
\end{tabular}
\end{table}

\begin{table}[t]
\setlength{\tabcolsep}{13pt}
\caption{Accuracies on classifications of gender, hand and fingers compared with those achieved by Shehu \cite{shehu2018detailed}.} 
\label{tab:genderhandfingers}
\begin{tabular}[t]{lcccc} \toprule
\multicolumn{1}{l}{} & \textbf{Gender}           & \textbf{Hand}             & \textbf{Fingers}          \\ \midrule \midrule
Shehu \cite{shehu2018detailed}       & 75.20\%          & 93.50\%          & 76.72\%          \\
\textbf{Our}         & \textbf{92.52\%} & \textbf{97.53\%} & \textbf{92.18\%} \\ \toprule
\end{tabular}
\end{table}






%
\section{Proposed Classification Solution}\label{sec:approach}

The main contribution of this paper is a \emph{single architecture/multi-task} solution that aims to solve all the classification tasks related to the "altered fingerprint" domain problem. In particular the combination of all the classifiers can be of great use to investigators in order to solve some of the historical examples that were previously described in Section \ref{sec:intro}. 

A Convolutional Neural Network (CNN), specifically the "Inception-v3" architecture, has been employed for all the tasks in order to extrapolate features from images and classify them. The extracted features are demonstrated to obtain a translation invariance, making this CNN architecture one of the best models for image classification tasks \cite{szegedy2016rethinking}.

The objective of this single architecture is to solve five different tasks: (i) discerning altered from authentic fingerprints, (ii) detecting the type of alteration, (iii) recognizing which finger or (iv) which hand the fingerprint in question comes from and, finally, (v) distinguishing fingerprints from male or female individuals.

\subsection{Neural Network design and parameters}

Starting from the idea of \cite{shehu2018detailed}, the CNN was trained from scratch only for the altered fingerprint classification task and for the remaining tasks only the final fully connected layers were fine-tuned thus creating the desired single/architecture feature. The CNN was trained with the following hyper-parameters: optimizer = Root Mean Square Propagation (RMSPROP), Learning Rate = 0.001, Epochs=25, Step per epoch=1000, Batch size = 32. Details on data will be given in the following Section.

The effectiveness of the employed technique has been demonstrated not only by results that will be presented in Section \ref{sec:results} but also in the empirical analysis of the activation maps as shown in Figure \ref{fig:activationMaps}.

\subsection{Data Preparation}\label{sec:dataprep}
The dataset used for training and test phases was the Sokoto Coventry Fingerprint Dataset (SOCOFing), developed by the Faculty of Engineering, Environment and Computing, Coventry University \cite{shehu2018sokoto}. SOCOFing is made up of 6,000 fingerprints belonging to 600 African subjects over the age of 18. Ten individual fingerprints were acquired from each individual and labelled with information regarding the gender of the individual they belonged to and the finger and hand from which they were acquired.
The acquisitions were made using the Hamster Plus (HSDU03P) and SecuGen SDU03P scanner sensors in the ".bmp" bitmap extension. Furthermore, using the STRANGE framework \cite{papi2016generation}, altered versions of the impressions were synthetically produced, providing three different types of alterations: obliteration, central rotation and Z-cut. This resulted in 17,934 impressions with slight alterations, 17,067 with medium alterations and 14,272 with a high level of alteration, for a total of 49,273 images. All images are grayscale and have a resolution of 96 x 103 pixels.

For the training of the CNN from scratch, the Inception-v3 was firstly trained and tested for the alteration binary test using: 16,388 footprints with central rotation, 17,866 obliterated footprints, 14,978 that have undergone a type alteration Z-cut and 6000 real. For each category, 50 \% of the images were used as Training-set, 20 \% as Validation-set and the remaining 30 \% as Test-set.

For all the other tasks, a fine-tuning training process of only the last fully connected layer was carried out on the training set by considering only the non-altered fingerprints. This was suggested by the activation maps themselves as shown in Figure \ref{fig:activationMaps} from which is possible to understand how the CNN concentrate on extracting features from altered zones, which is not the case for tasks like gender or hand recognition. In this way the fully connected layers focus on the non-altered (predicted) parts of the images for these tasks.

\section{Results}\label{sec:results}

The Inception V3, trained as described in the previous Sections, and validated for hyper-parameters was finally tested for all the five tasks. In this Section, obtained results will be presented and commented w.r.t. best results in the state of the art.

\subsection{Fakeness Detection}\label{sec:fakeness}
This is a binary classification problem that has the objective to discriminate between authentic and altered fingerprint images. Results are shown in Table \ref{tab:fakeness}. The accuracy obtained is of  98.21 \%. Indeed Tabassi et al. in \cite{tabassi2018altered} obtained a better accuracy of 99.24 \% by means of a different dataset. Moreover, the precision achieved is of 99.97\%; therefore, the consequent False Positive Rate is much lower than \cite{tabassi2018altered} and this is particularly relevant for fingerprint-based security systems. Finally, this is just the first analysis to be carried out. If other classifiers detects anomalies (gender, finger or hand-side mismatching), this could raise an alarm for further investigations.

\subsection{Alterations Detection}\label{sec:alteration}
Alteration type detection is a ternary classification problem between the three alteration types described in this paper with, in addiction, fourth class representing non-altered (real) fingerprints. In this test results will challenge the state of the art obtained by Shehu et al. \cite{shehu2018detection} 
Results obtained are shown in Table \ref{tab:alteration} with an accuracy of 98.46\% that overall is comparable with the accuracy obtained by Shehu of 98.55\%. However again, precision and recall are better than those evaluated by Shehu by 1\% on average for both metrics. Further demonstrating the more robust characteristic of the approach. And this will be clearer in the next sections were, using the same architecture, three further classification tasks will be carried out.

\subsection{Gender Classification}\label{sec:gender}
Gender classification is a binary classification problem. Here we compare the results with those obtained by Shehu et al. in \cite{shehu2018detailed} . Given the sample imbalances between male and female, a random undersampling technique was employed in order to give, for each batch, equal amount of samples between male and female.
Results are shown in Table \ref{tab:genderhandfingers}. The accuracy obtained is of 92.52\%, which is far more better than 75.20\% obtained by Shehu et al. demonstrating the robustness of the technique.

\subsection{Finger and Hand Classification}\label{sec:finger}
Finally, classifiers for finger and hand-side tasks were tested. For this test, results are compared with those obtained by Shehu et al. in \cite{shehu2018detailed}, which is the best in the state of the art. The first is a 5-class problem while the second is a binary classification problem. Results are shown in Table \ref{tab:genderhandfingers}. The accuracy obtained is, respectively, of 92.18 \% for finger prediction, which is much lower than 76.72\% achieved by Shehu et al., and of 97.53\% for hands prediction against a 93.5 \% for Shehu et al. The improvement in this task demonstrates that the proposed approach has reached a better generalization and, probably, can maintain these levels of performance in large scale usage.

\section{Conclusions}\label{sec:conclusion}
Fingerprints are the most numerous and valuable source of information in crime scene investigation. But criminals have many methods for altering them introducing or removing the dermatoglyph crests which constitute the biometric information. In this paper, a \emph{single architecture/multiple task} Neural Network solution based on Inception-v3 has been trained and tested for five different tasks, achieving results better than state of the art. This solution showed robustness and generalizability and a little bit of explainability, since the activation maps detected alterations.

%

\bibliographystyle{IEEEbib}
\bibliography{refs}



%








\end{document}